\documentclass[letterpaper, 10 pt, conference]{ieeeconf}  

\IEEEoverridecommandlockouts                              

\usepackage{graphicx} 
\usepackage{mathptmx} 
\usepackage{amsmath} 
\usepackage{amssymb}  
\usepackage{cleveref}
\usepackage{comment}
\crefname{figure}{Fig.}{Fig.}
\crefname{table}{Table.}{Table.}

\newenvironment{nolabel_caption}{
    \footnotesize
    \baselineskip = 3.5mm
    \flushleft
}

\title{\LARGE \bf
Naming Objects for Vision-and-Language Manipulation
}

\author{
Tokuhiro Nishikawa$^{1*}$,
Kazumi Aoyama$^{1*}$,
Shunichi Sekiguchi$^{1*}$,
Takayoshi Takayanagi$^{1}$,
Jianing Wu$^{1}$, \\
Yu Ishihara$^{1}$,
Tamaki Kojima$^{1}$, 
and Jerry Jun Yokono$^{1}$
\thanks{*These authors contributed equally to this work.}
\thanks{$^{1}$Sony Group Corporation, 1-7-1 Konan Minato-ku, Tokyo, 108-0075, Japan. Corresponding: {\tt\small tokuhiro.nishikawa@sony.com}}%
}

\begin{document}

\maketitle
\thispagestyle{empty}
\pagestyle{empty}

\begin{abstract}
Robot manipulation tasks by natural language instructions need common understanding of the target object between human and the robot. However, the instructions often have an interpretation ambiguity, because the instruction lacks important information, or does not express the target object correctly to complete the task. 
To solve this ambiguity problem, we hypothesize that "naming" the target objects in advance will reduce the ambiguity of natural language instructions. 
We propose a robot system and method that incorporates naming with appearance of the objects in advance, so that in the later manipulation task, instruction can be performed with its unique name to disambiguate the objects easily.
To demonstrate the effectiveness of our approach, we build a system that can memorize the target objects, and show that naming the objects facilitates detection of the target objects and improves the success rate of manipulation instructions.
With this method, the success rate of object manipulation task increases by 31\% in ambiguous instructions.
\end{abstract}
\section{INTRODUCTION}
\label{sec:INTRODUCTION}

Toward the home service robots that can perform multiple tasks, natural language interface is important, as it is one of the most easiest way humans can communicate freely without the need of special training. Therefore, robot manipulation with natural language interface is widely studied \cite{min2021film, misra2016tell, jiang2022vima,shridhar2022cliport,hatori2018interactively}.

Manipulation, such as "pick and place", is the most important tasks for the home service robots. In order to perform these tasks by natural language instructions, there needs common understanding between the human and the robot. However, the instructions with natural language often lacks important information, or does not express the target object correctly to complete the task. 

\begin{figure}[t!]
    \vspace{2mm}
    \begin{center}
        \begin{minipage}[b]{0.49\hsize}
            \begin{center}
                {\footnotesize Instruction: "Pick the bottle up."}\\
                \vspace{1mm}
                \includegraphics[height=3.2cm]{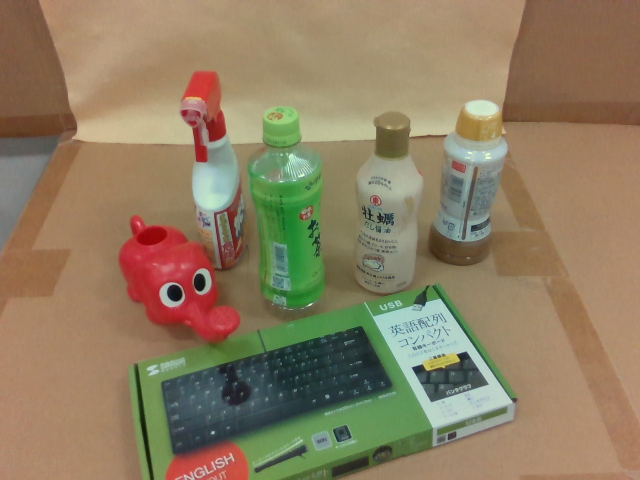}
            \end{center}
        \end{minipage}
        \begin{minipage}[b]{0.49\hsize}
            \begin{center}
                {\footnotesize Instruction: "Pick up the cat."}\\
                \vspace{1mm}
                \includegraphics[height=3.2cm]{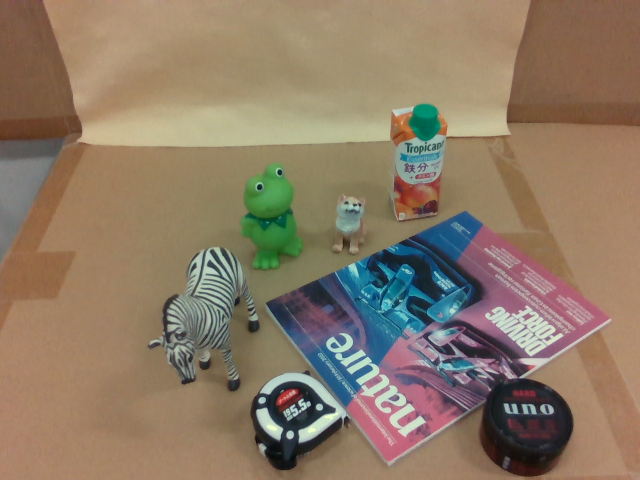}
            \end{center}
        \end{minipage}
    \end{center}
    \vspace{-4mm}
    \caption{Examples of natural language instruction including ambiguous expression : 
    \textbf{Left:} Instruction with multiple candidates for the target objects. (Three bottles in sight.) \textbf{Right:} The expression of the target object is incorrect. (There is no cat in sight but the dog is.)}
    \label{fig:ambiguous_situation}
    \vspace{-4mm}
\end{figure}

In our research, we found that 17\% of human-created object manipulation instructions were containing ambiguity in target object expression, even though the annotators were asked to make object manipulation instruction detail enough to make the target object identifiable by the robot (Details are in \cref{sec:EXPERIMENTAL EVALUATION}).
In this paper, we define the ambiguity of manipulation instruction as follows: 1) There are multiple candidates for the target object. 2) The referring expression of the target object is incorrect. Examples are shown in \cref{fig:ambiguous_situation}.
Besides, Hatori et al. \cite{hatori2018interactively} reported that 21\% of the object manipulation instructions are judged as it has multiple target object candidates, and treated as ambiguous instruction. 
Hence, object manipulation instructions using natural language are prone to contain ambiguous expressions, and as a result, the robot cannot identify the target object, which the manipulation instructions are not executed as intended.

To solve these ambiguity problems, several methods that a robot gives extra feedback to a operator \cite{ hatori2018interactively, whitney2017reducing, yang2022interactive} are proposed. 
However, these methods require additional interaction steps every time to clarify the instruction if there are ambiguities, which we believe we can omit by \textit{naming} it.

Thus, we hypothesize that "naming" the target objects in advance will reduce the ambiguity of natural language instructions. 
Once the name of the object is shared with the robot and the human,
the target object is uniquely identified by the name.
As a result, additional interactions are not required unlike the previous methods \cite{ hatori2018interactively, whitney2017reducing, yang2022interactive}.
We propose a robot system and method that incorporates naming with appearance of the objects in advance, so that in the later manipulation task,  instruction can be performed with its unique name to disambiguate the objects easily.
To demonstrate the effectiveness of our approach, we build a robotic system that can memorize the appearance and name of the target objects, and show that naming the objects facilitates detection of the target objects and improves the success rate of manipulation instructions. With this method, the success rate of object manipulation task increased by 31\% in ambiguous instructions.

To summarize, our paper's contributions are as follows:
\begin{itemize}
    \item  A proposal that naming the objects with its appearance in advance helps to reduce the ambiguity of target objects in object manipulation instructions with natural language.
    \item  A proposal for a system that can memorize names with its appearance of the objects and recall them.
    \item  Through experiments in simulation and with real robot system to prove above proposal are effective in robot manipulation task with natural language.
\end{itemize}

\label{sec:RELATED WORKS}
\section{RELATED WORKS}

A number of studies have been conducted to enable the robots to follow the natural language instructions 
\cite{
min2021film,
misra2016tell,
jiang2022vima,
shridhar2022cliport, 
hatori2018interactively,
yang2022interactive,
jang2022bc,
ishikawa2021target}.
The manipulation by natural language instruction is one of hot topics in this field.
For example, Hatori et al. \cite{hatori2018interactively} built an interactive system to which user can use unconstrained spoken language instructions to operate a object picking task.
CLIPort \cite{shridhar2022cliport} can solve variety of language instructed tasks from packing objects to folding cloths.
 
Referring expression comprehension (REC) \cite{kazemzadeh2014referitgame} is closely related to natural language instructed robot manipulation.
REC aims to localize the target object in an image described by a referring expression.
In vision-and-language manipulation, it is also necessary to locate the object in the instructions.
MDETR \cite{mdetr_2021_ICCV} and GLIP \cite{li2021grounded} are capable of estimating the position of objects referred by natural language, based on large models trained on large data. 

During the robot manipulation and in its grounding process of instructions, the system must recognize the target object. 
However, there often happens to be a case, that human refers the target object with ambiguity, and the system cannot ground it properly. 
When there are ambiguities in human language instructions, the robot needs to resolve the ambiguity using information other than the instruction, to specify the target object.
As an illustration, when the instruction from a person is "Pass me my bottle.", in a situation where there are multiple bottles available, the robot may have several options to resolve which is "my bottle": Ask that person back, recognize the gaze or the pointing-gesture of the person, or refer to the past memory.
Some researches have proposed a language-based feedback to robots to disambiguate instructions \cite{hatori2018interactively, yang2022interactive, whitney2017reducing}.
Whitney et al. \cite{whitney2017reducing} showed that when the robot cannot determine one object, a simple confirmation question, i.e., “This one?” is useful.
Hatori et al. \cite{hatori2018interactively} showed that asking for a rephrase of the instruction and reinterpreting them improves the recognition performance of the referred-to object.
Yang et al. \cite{yang2022interactive} demonstrated that using object attributes is useful in disambiguation, by developing an grasping system capable of resolving ambiguities via dialogue.
However, although language feedback is an effective tool for disambiguation, the additional interactions on every ambiguous instruction might degrade the usability of the robot.

One way humans handle ambiguities in daily communication is lexical entrainment.
Lexical entrainment is a psychological phenomenon of people tending to adopt the terms of their conversation partner.
Iio et al. have shown that lexical entrainment also occurs in human-robot interaction \cite{iio2015lexical}.
For instance, when a robot refers to an object by its color, a human also refers to the object by its color, and when a robot calls an object by its name, a human also calls the object by its name.
They argued that facilitating lexical entrainment between human and robot, i.e., calling the object by a common term, would lead to better recognition performance of the referred-to object by the robot, but they did not conduct experiments to confirm that hypothesis.

As for related works that proposed naming object concept, Nakamura et al. \cite{nakamura2012learning} pointed out that objects have names that are used only within a family, and that it is desirable for humans and robots to be able to understand and refer to objects by the same name.
They proposed a method to learn novel objects and their names from audio-visual input and evaluated the method with a task adopted from the RoboCup@Home league \cite{robocuphome}.
However, they did not clearly show the advantage of using names for robot tasks.
Jiang et al. \cite{jiang2022vima} trained a large-scale vision and language model and showed that the named objects can be grasped by instructions combined with images and language.  
However, they did not show the benefits of using the names. 
There are some benchmarks on manipulation by language instructions \cite{zheng2022vlmbench,mees2022calvin,shridhar2020alfred}.
However, these benchmarks do not focus on resolving human language ambiguity.

Thus, there is no literature that shows the benefit of sharing the object names between human and robot.
In this paper, we propose a novel system that can memorize the names of the objects, and show that naming the objects facilitates detection of the target objects and improves the success rate of manipulation instructions.

\section{Naming Objects for Vision-and-Language Manipulation}
\label{sec:Naming Objects for Vision-and-Language Manipulation}

\begin{figure*}[!htb]
    \centering
    \includegraphics[scale=0.24]{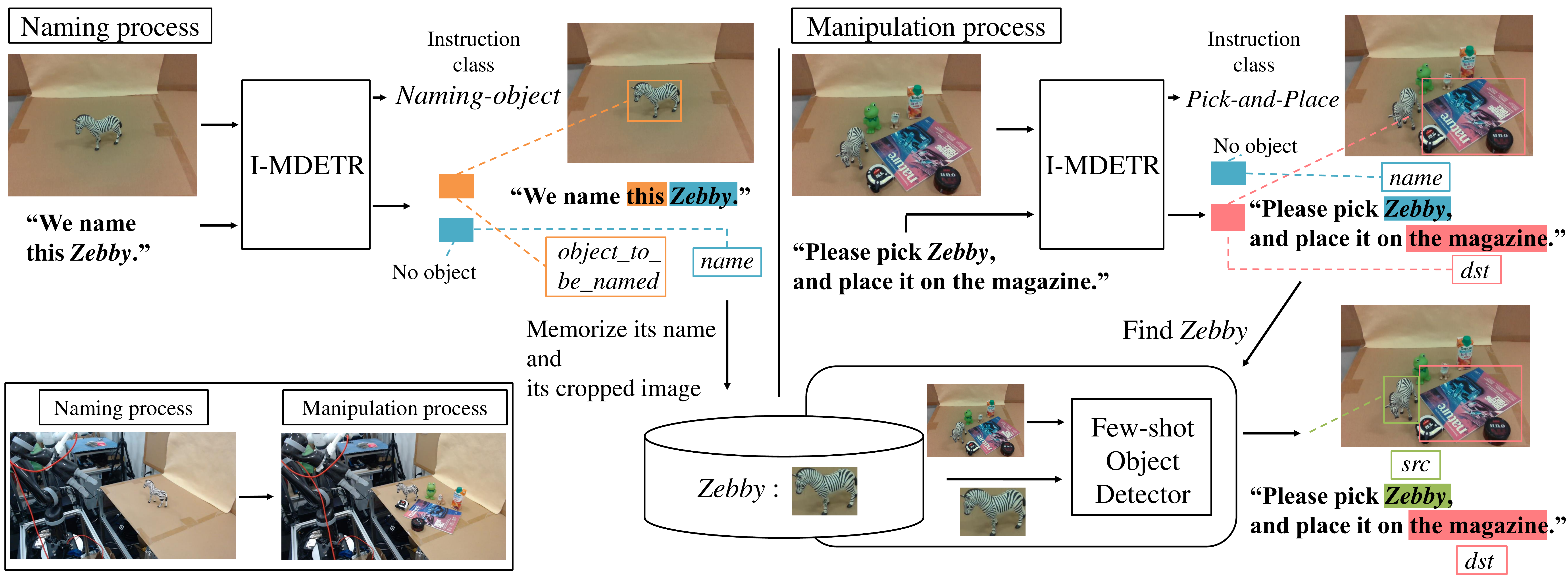}
    \vspace{-3mm}
    \caption{We propose a system for naming the target object in advance and using that name in pick-and-place instructions. 
    In the naming process, by using I-MDETR, we first detect the object to be named in the scene and its name in the instruction, and then store its image and its name in the memory as shown in the left side. After the naming process, the system receives a pick-and-place instruction including the name and finds the object to pick and the placing position by using I-MDETR and a few-shot object detector as shown in the right side.
    }
    \label{fig:proposed_system}
    \vspace{-1mm}
\end{figure*}

\begin{figure}[!htb]
    \centering
    \includegraphics[scale=0.24]{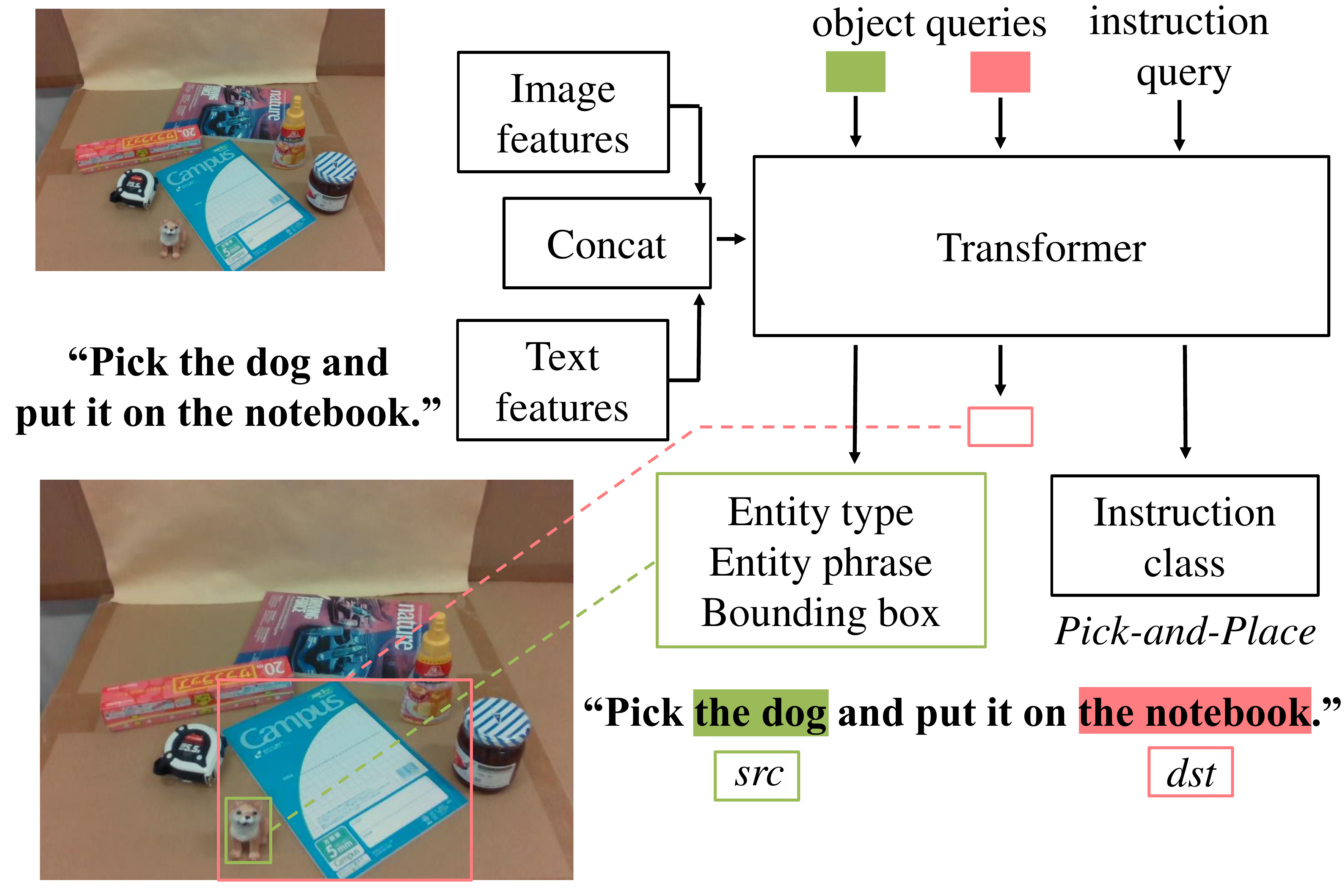}
    \vspace{-5mm}
    \caption{Instruction-MDETR (I-MDETR) overview. I-MDETR is an extended version of MDETR \cite{mdetr_2021_ICCV}. It predicts instruction class, entity type and phrases corresponding to it as well as the bounding boxes of the objects.}
    \label{fig:instruction_mdetr}
    \vspace{-2mm}
\end{figure}

The goal of our research is to build a robot system that performs pick-and-place with user-provided natural language instruction.
Our system works in the following steps.
\begin{enumerate}
\item Receive a natural language instruction (text) and scene information (RGB-D image).\label{step1}
\item Predict the object to pick and the position to place.\label{step2}
\item Manipulate the object using predicted information.\label{step3}
\end{enumerate}
However, the ambiguity of natural language instruction makes difficult to identify the target object in \ref{step2}).
If the system fails predicting the object to pick and the placing position, the system will not work as the user intended. 
Hence, to reduce such ambiguity and improve object manipulation performance, we propose naming the object in advance, and use that name in the instructions to correctly identify the picking objects and the placing position.
By naming the object in advance, the ambiguity of language instruction is mitigated and the system can achieve the pick-and-place task.

Proposed algorithm consists of two main processes.
\begin{itemize}
    \item \textbf{Naming process}
        \begin{enumerate}
            \item Receive a naming instruction and an image of the scene.
            \item Detect the object to be named in the scene and its name in the instruction.
            \item Store its image and its name in the memory.
        \end{enumerate}
    \item \textbf{Manipulation process}
        \begin{enumerate}
            \item Receive a pick-and-place instruction and an image of the scene.
            \item Find the object to pick and the placing position.
            \item Manipulate the object according to the information extracted in 2).
        \end{enumerate}
\end{itemize}
To realize the step 2) of the naming and the manipulation process, we introduce Instruction MDETR (I-MDETR), a MDETR \cite{mdetr_2021_ICCV}-based object detector extended for language instructions.
In addition, for the step 2) of the manipulation process, we also implement a few-shot object detector \cite{arcface, cui2022mixformer} to find the named object in the instruction from objects detected by I-MDETR with memorized images.
In the following sub-sections, we will describe I-MDETR in detail and explain the algorithm flow of the naming and the manipulation process.

\subsection{Instruction MDETR (I-MDETR)}
I-MDETR is an extension of MDETR \cite{mdetr_2021_ICCV}, a method proposed for Referring Expression Comprehension (REC) tasks that predicts bounding boxes of the objects in an image with their grounding in text phrase.  
Compared to MDETR, I-MDETR also predicts entity type which denotes the role of the phrase in the instruction. \cref{fig:instruction_mdetr} shows the architecture of I-MDETR.
We define the followings as entity types for our task setting.
\begin{itemize}
    \item \textbf{name}: Name of the object.
    \item \textbf{object\_to\_be\_named}: The object to be named.
    \item \textbf{src}: Target object of picking.
    \item \textbf{dst}: Target object that indicates the place position.
\end{itemize}
Please note that it is possible to add other entity types to cover different task settings.
Additionally, to classify the objective of the instruction itself, I-MDETR also predicts instruction class.
The following instruction classes are defined for our task.
\begin{itemize}
    \item \textbf{naming-object}: Instruction for naming.
    \item \textbf{pick-and-place}: Instruction for pick-and-place.
    \item \textbf{instruction-not-supported}: Instruction not supported.
\end{itemize}
\textbf{instruction-not-supported} are introduced to properly handle instructions that does not relate to our manipulation task.
Note that as well as entity type, other instruction classes can be added to cover different task settings.

\subsection{Naming Process}

When I-MDETR predicts \textbf{naming-object} as instruction class, we first extract the phrase of entity type \textbf{name} and \textbf{object\_to\_be\_named} in the instruction from the output of I-MDETR. 
Specifically, we search for a bounding box with \textbf{object\_to\_be\_named} phrase and check its entity type for the name of the object in the bounding box (See also \cref{fig:proposed_system}). If the system fails in finding the target object and its name, it will end processing the request at this point. Otherwise, when the system finds the object and its name, it saves the object in the bounding box as an image tagged with the provided name in the instruction. By tagging the image with the name of the object, we can find the named object by matching the object in the bounding box with few-shot detectors during the manipulation process.

\subsection{Manipulation Process}
When I-MDETR predicts \textbf{pick-and-place} as instruction class, the system performs the manipulation in the following procedure.
First, using the bounding boxes and phrases (and their entity types) detected by I-MDETR, the system searches for the target object to pick up and target placing position. The target object to pick up is chosen by searching for the entity type \textbf{src} or \textbf{name} in the instruction. When \textbf{src} is found in the instruction, the corresponding bounding box of an object in the image is used as target object to pick. However, when entity type \textbf{name} is found in the instruction, the bounding box output by I-MDETR could be incorrect. This is because I-MDETR is not trained with the name of the object. Therefore, we make use of few-shot object detector \cite{arcface, cui2022mixformer} to find the object with given name in the instruction (See Appendix \ref{appendix:few-shot} for the details of few-shot object detector). Concretely, we use the few-shot detector to find the object in the scene that matches with saved images and use the most matched object as target object. When multiple entity type \textbf{src} or \textbf{name} are found in the instruction, we select the most probable entity type according to the confidence score output by I-MDETR. The target placing position is also chosen similarly using the \textbf{dst} or \textbf{name} in the instruction.
After the target object and placing position are determined, we run the pick-and-place operation by combining predefined manipulation skill set of the manipulator. 

\subsection{Training of I-MDETR and Other Components}
For the training of I-MDETR, we implemented an automatic training dataset generation flow with a physics engine, Mujoco \cite{todorov2012mujoco}.
60 common household objects are 3D-scanned and arranged on a table in Mujoco simulation environment.
Instruction texts and annotations are automatically generated by scripts using human-annotated expression templates.
Based on the MDETR weights pre-trained with RefCOCOg \cite{mao2016generation}, we finetuned I-MDETR model with our generated dataset.
RefCOCOg is a widely used dataset for REC in vision-and-language field.
The details of the training dataset for I-MDETR are provided in Appendix \ref{appendix:data_gen_IMDETR}.
We also attached detailed information of our pick-and-place method to Appendix \ref{appendix:VGN}.

\section{EXPERIMENTAL EVALUATION}
\label{sec:EXPERIMENTAL EVALUATION}

\begin{figure}[!tb]
    \centering
    \includegraphics[scale=0.24]{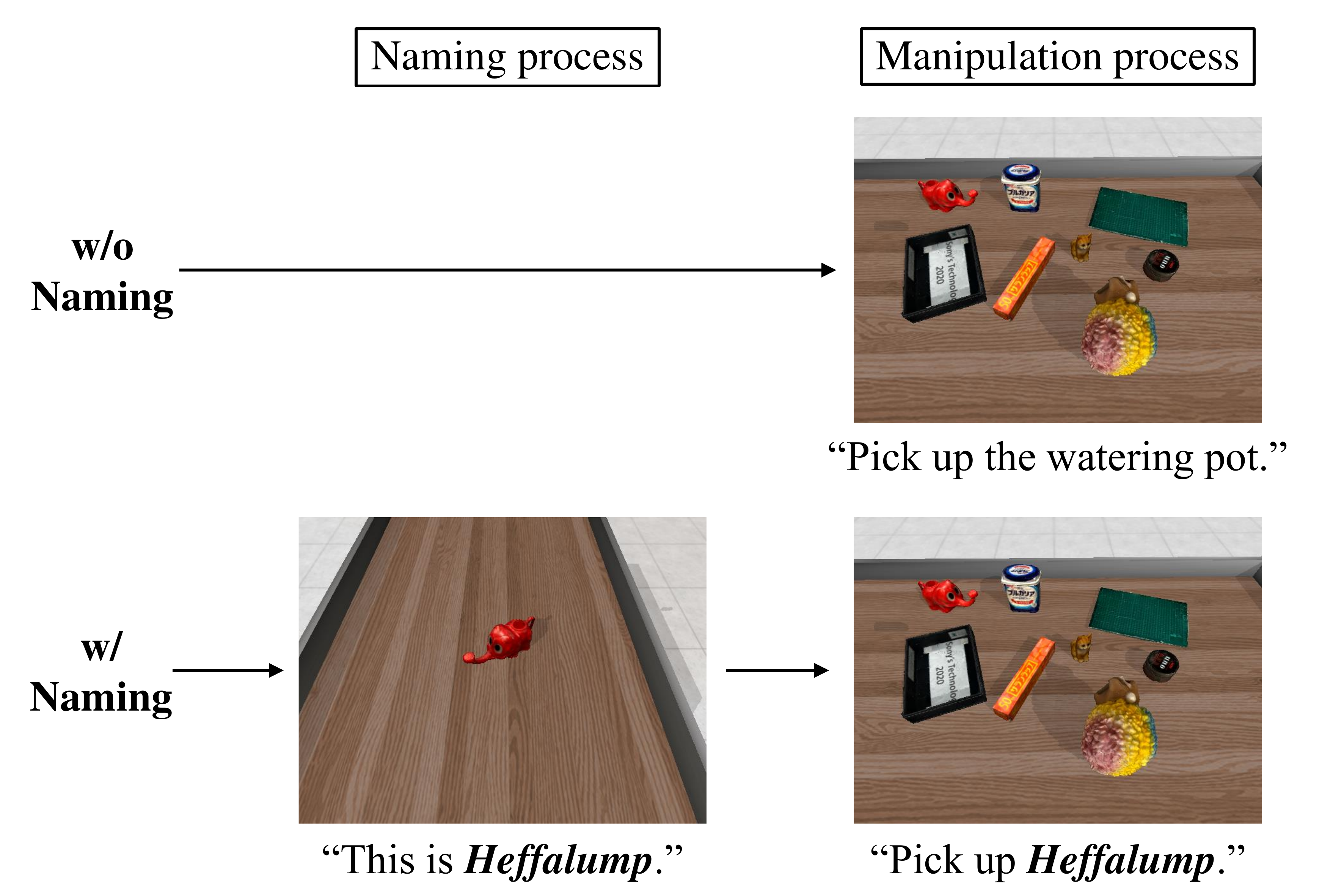}
    \vspace{-6mm}
    \caption{Experiment procedure. In without naming condition({\bf w/o Naming}), we simply instruct the robot to do a pick-and-place instruction. In with naming condition({\bf w/ Naming}), we first tell the robot to memorize the object by its name, then instruct to do pick-and-place with the name.}
    \label{fig:experiment_sim}
    \vspace{-2mm}
\end{figure}

Our experiments aim to answer the following question: Does naming the objects with its appearance in advance help reduce the ambiguity of target objects in natural language instructions and improve the success rate of manipulation?
Our experiment procedure is shown in \cref{fig:experiment_sim}.
We compare the manipulation performance without and with naming.
\begin{itemize}
    \item{\bf w/o Naming}: Robot is simply instructed to do a pick-and-place task, with scene image and instruction text given to the robot system.
    \item{\bf w/ Naming}: First, robot is instructed to memorize an object by its name, then instructed to operate a pick-and-place task with the name.
\end{itemize}
In w/ naming, when naming a object, the user shows only one object to the robot and tells its name.
In naming, after I-MDETR recognize the object (obj-to-be-named), it stores a cropped image of the object in its memory.
In our experiment, we assume the robot can look at the object multiple times from different views.
In the experiment, 4 images of different view angles are saved in its memory.

We conducted the experiments with human-generated language instructions in both simulation and real-robot.
The following sections give the details of human instruction collection, evaluation metrics, and results.

\subsection{Human Instruction Collection}
\label{subsection:human_instruction_collection}
We used the same 60 objects that we used for the training dataset to generate evaluation scenes.
We randomly selected 6 to 8 objects among 60 objects and placed them on a desk.
Camera positions and postures were selected in the same way as the training dataset generation.
Total 20 scenes were generated for the evaluation.

We asked 15 volunteers to annotate an instruction for each scene and a total of 300 instructions are collected (15 instructions per scene).
The volunteers were requested to follow the directions below.
\begin{itemize}
\item Make pick-and-place instructions in natural language.
\item Do not write an instruction that requires to pick and/or to place multiple objects.
\item No instructions to move the object off the table, such as "Pick the can and put it in the refrigerator."
\item The target object should be identifiable uniquely.
\end{itemize}

There are two reasons for collecting evaluation instructions from human volunteers in this way.
One is to investigate how often humans give ambiguous instructions, and the other is to investigate what kind of ambiguous instructions they make.
Prior researches have shown that even when directed to uniquely refer to an object, humans sometimes give ambiguous instructions \cite{hatori2018interactively, kazemzadeh2014referitgame, mao2016generation}.

After the collection, we asked an experienced human annotator to check all 300 instructions to see if how many instructions have ambiguity in identifying target objects.
As a result, 51 instructions, which is 17\% of the total, were found to be ambiguous.
To be precise, there were 31 cases where the object was not uniquely identified from the instruction (i.e., there were multiple candidates), and 24 cases where the object was not exactly correctly referenced due to misunderstanding of target object.
We confirmed that some instructions are still ambiguous when checked by the third person (i.e., the experienced annotator) strictly, even if volunteers are asked to annotate pick-and-place instructions with which the operator(i.e., robot) can uniquely identify the target object without interpretation ambiguity.
In other words, it is difficult to eliminate ambiguity from human instructions, so unless the robot can resolve these ambiguities, the robot cannot successfully pick-and-place objects.

Annotation of instruction class, bounding boxes, entity phrases, and entity types were also done manually by the annotator.
Even when the target object in the instruction is ambiguous, there is one object that the volunteer intends to manipulate.
Therefore, for these ambiguous instructions, the annotator guessed the volunteer's intent and selected a target.

Even though we had expected volunteers to write pick-and-place instructions, 
some of collected instructions were actually impossible for our robot to carry out, e.g., "Balance the notebook on the bottle."
The instruction class of such instruction is manually replaced with "{\bf instruction-not-supported}", because we consider that the robot needs to be capable of ignoring them, especially in real use case.
These human-annotated instructions created by the above procedure are used for the w/o naming condition of the experiment.

Based on the evaluation data for w/o naming, we generated evaluation data for w/ naming condition of our experiment.
We simply replace referring expressions in the instructions with names.
The name for the object is labeled manually by a volunteer by looking at the object.
Naming instructions are generated by randomly selecting from pre-defined naming sentence templates(e.g., "The name of that toy is A" or "I call it A").

\subsection{Evaluation Metrics}
We define the following evaluation metrics.
\begin{itemize}
    \item {\bf ICR}: Instruction classification success rate.
          The rate of instructions I-MDETR correctly predicted the instruction class out of the total instructions.
    \item {\bf PR}: Phrase extraction success rate.
          The rate of instructions I-MDETR correctly extracts the phrases of target objects out of the total instructions.
          Depending on the entity type of the object, it is either the name or the referring expression of the object.
    \item {\bf BR}: Bounding box detection success rate.
          The rate of instructions I-MDETR successfully detects the target object region in the input image out of the total instructions.
          Success if the IoU of detected bounding box and ground truth is more than 0.5.
    \item {\bf SR}: Process success rate.
          The rate of instructions the system successfully complete the given process (naming or manipulation) out of the total instructions. 
          In w/ naming, to successfully complete manipulation process, the robot has to succeed naming process in advance.
\end{itemize}

\subsection{Simulation Experiment Results}
\begin{table*}[!bt]
    \caption{Simulation Experiment Results}
    \begin{center}
        \begin{tabular}{llccccc}
            \hline
                       &                  & Naming process   & \multicolumn{4}{c}{Manipulation process}                          \\
                       &                  & \bf{SR} [\%]     & \bf{ICR} [\%]  & \bf{PR} [\%]   & \bf{BR} [\%]   & \bf{SR} [\%]   \\
            \hline
            w/o Naming & all instructions & -                & 95.3 (286/300) & 72.0 (216/300) & 57.0 (171/300) & 57.0 (171/300) \\
                       & unambiguous      & -                & 95.2 (237/249) & 71.5 (178/249) & 63.0 (157/249) & 62.2 (155/249) \\
                       & ambiguous        & -                & 96.1 (49 / 51) & 74.5 (38 / 51) & 27.5 (14 / 51) & 31.4 (16 / 51) \\
            \hline
            w/ Naming  & all instructions & 98.3 (295/300)   & 94.0 (282/300) & 83.7 (251/300) & 69.3 (208/300) & 69.7 (209/300) \\
                       & unambiguous      & 98.8 (246/249)   & 94.4 (235/249) & 85.5 (213/249) & 70.1 (176/249) & 71.1 (177/249) \\
                       & ambiguous        & 96.1 (49 / 51)   & 92.2 (47 / 51) & 74.5 (38 / 51) & 62.7 (32 / 51) & 62.7 (32 / 51) \\
            \hline
        \end{tabular}
    \end{center}
    \label{table:sim_result_all}
\end{table*}

As mentioned in Section \ref{subsection:human_instruction_collection}, there are 51 ambiguous instructions and 249 unambiguous instructions, for a total of 300.
With these instructions, we conducted both w/o naming and w/ naming experiments.
\cref{table:sim_result_all} shows the results.

From \cref{table:sim_result_all} we can confirm that {\bf SR} of manipulation process for all 300 instructions in w/ naming was 69.7\%, which is a 12.7\% improvement over the 57.0\% in w/o naming.
Focusing on 51 ambiguous instructions, {\bf SR} of Manipulation process was 62.7\% in w/ naming, while it is 31.4\% in w/o naming.
This is almost a 2-fold performance improvement.
These results indicate that naming objects effectively mitigates the ambiguity of instructions.
Looking into the results in detail, from the fact that {\bf PR} for ambiguous instructions are same for both w/ and w/o naming, the 35.2\% of {\bf BR} improvement contributed the {\bf SR} improvement.
The reason of {\bf BR} improvement is considered to be the reduction of ambiguity by naming process.
For ambiguous instructions, the entity phrases of the target object can be extracted from the instruction text, but the bounding box detection often fails because of the ambiguity.
In w/o naming, I-MDETR should detect the object based on the given ambiguous instruction.
Conversely, in w/ naming the few-shot object detector detects the target by using the image of the object that is memorized beforehand in the naming process.
These results indicate that when the target object is ambiguously referred in the instruction, it becomes difficult to find the target object only by the referring expression.

Additionally, for 249 unambiguous instructions, {\bf SR} of manipulation process also improved by 8.9\% by naming.
The {\bf PR} and {\bf BR} are improved by 14.0\% and 7.1\% respectively.
The improvement of {\bf PR} implies that it is easier to extract names than to extract the referring expressions.
As a result of the improvement of {\bf PR}, {\bf BR} is also improved.
This is because, w/ naming, the system recalls the target object's memorized image by the extracted name and runs the few-shot object detector. 
Consequently, it seems naming is effective not only in resolving ambiguity, but also in making the target object detection easier.

\begin{figure*}[tb!]
    \begin{center}
        \begin{minipage}[t]{0.23\hsize}
            \begin{center}
                \includegraphics[height=3cm]{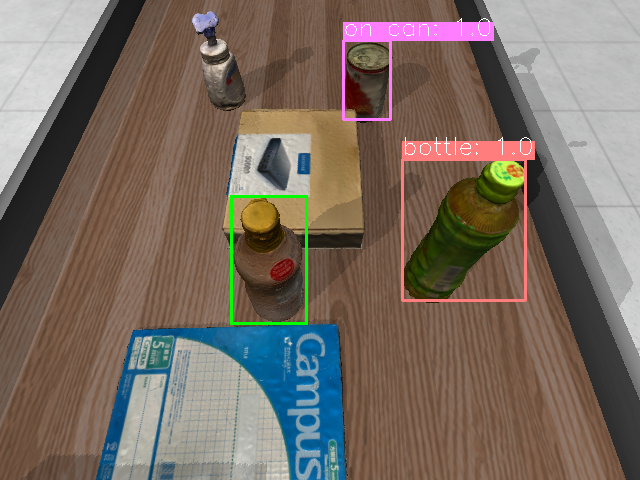}\\
                \vspace{-2mm}
                \begin{nolabel_caption}
                (a) {\bf "Pick up bottle and place on can."} There are multiple bottles. The intended target is the brown bottle in the center. \bf{ambiguous, failure}.
                \end{nolabel_caption}
            \end{center}
        \end{minipage}
        \begin{minipage}[t]{0.23\hsize}
            \begin{center}
                \includegraphics[height=3cm]{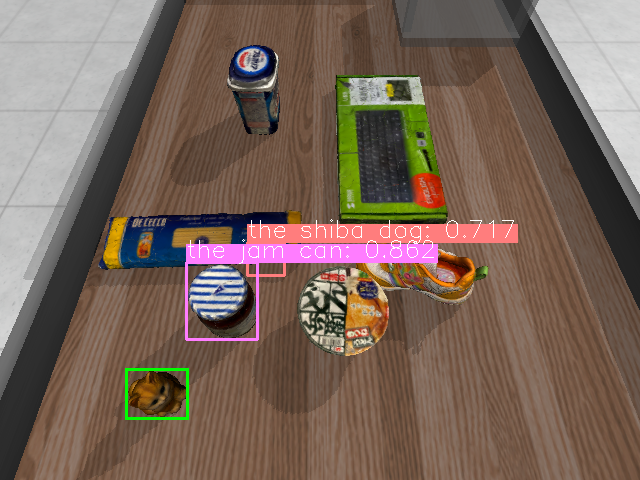}\\
                \vspace{-2mm}
                \begin{nolabel_caption}
                (b) {\bf "Pick up the shiba dog and put it on top of the jam can."} There is no dog but a cat, and detection of "shiba dog" failed. {\bf ambiguous, failure}.
                \end{nolabel_caption}
            \end{center}
        \end{minipage}
        \begin{minipage}[t]{0.23\hsize}
            \begin{center}
                \includegraphics[height=3cm]{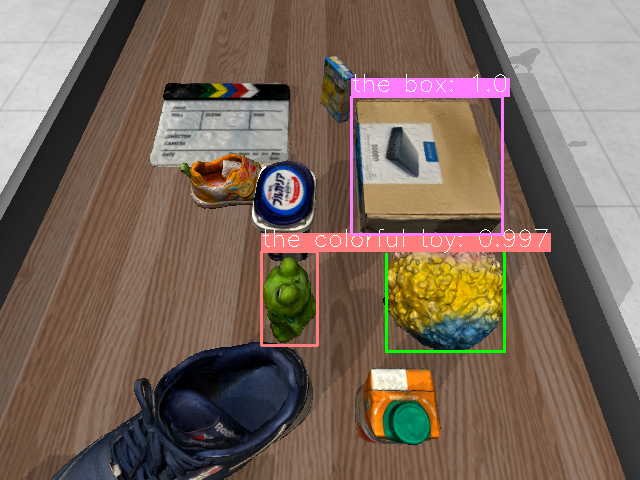}\\
                \vspace{-2mm}
                \begin{nolabel_caption}
                (c) {\bf "Pick up the colorful toy and place it on the box."} The word "colorful" seems subjective. The robot picked the green frog toy but the intended target was another toy. {\bf unambiguous, failure}.
                \end{nolabel_caption}
            \end{center}
        \end{minipage}
        \begin{minipage}[t]{0.23\hsize}
            \begin{center}
               \includegraphics[height=3cm]{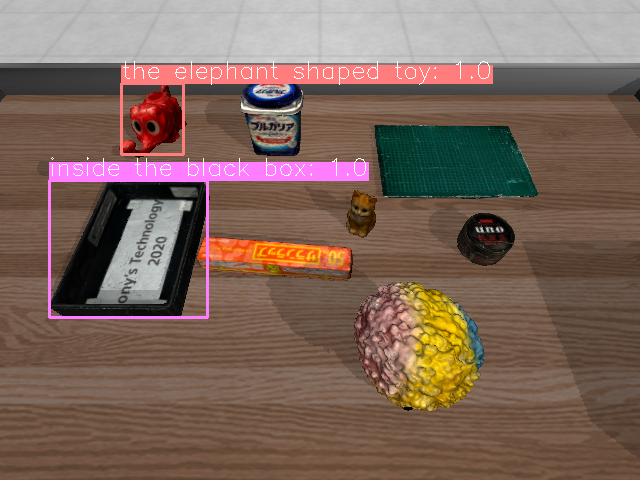}\\
               \vspace{-2mm}
               \begin{nolabel_caption}
               (d) {\bf "Pick the elephant shaped toy and put it inside the black box."} The elephant shaped object is a watering pot and whether it is a toy or not is a subjective judgment. {\bf unambiguous, success}.
               \end{nolabel_caption}
            \end{center}
        \end{minipage}
    \end{center}
    \vspace{-1mm}
    \caption{{\bf w/o Naming} examples. Green bounding box is ground truth. Pink and orange bounding boxes are predicted {\bf src} and {\bf dst}.}
   \label{fig:result_wthout_naming}     
\end{figure*}

\begin{figure*}[tb!]
    \begin{center}
        \begin{minipage}[t]{0.23\hsize}
            \begin{center}
                \includegraphics[height=3cm]{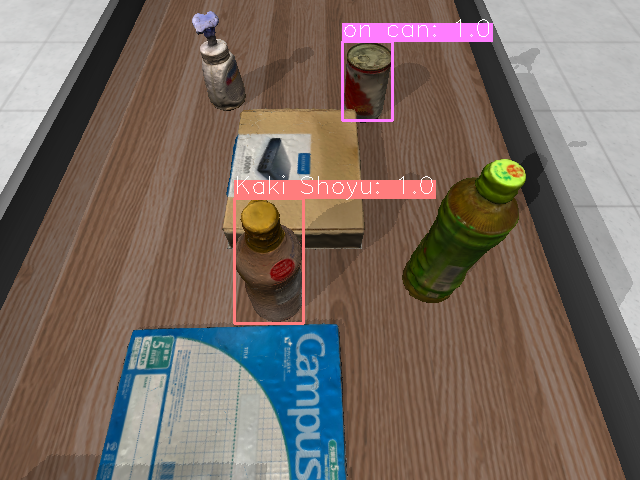}\\
                \vspace{-2mm}
                \begin{nolabel_caption}
                (a) {\bf "Pick up \textbf{\textit{Kaki Shoyu}} and place on can."} The original instruction was {\bf "pick up bottle and place on can."} and there are multiple bottles. Thanks to naming, there is no misunderstanding between human and robot this time. \bf{ambiguous, success}.
                \end{nolabel_caption}
            \end{center}
        \end{minipage}
        \begin{minipage}[t]{0.23\hsize}
            \begin{center}
                \includegraphics[height=3cm]{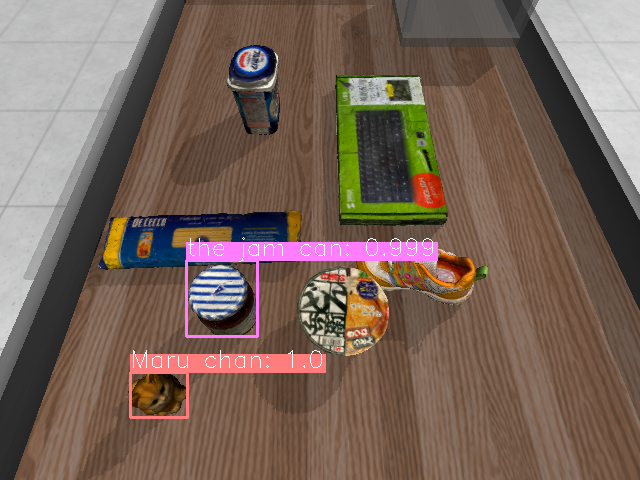}\\
                \vspace{-2mm}
                \begin{nolabel_caption}
                (b) {\bf "Pick up \textbf{\textit{Maru chan}} and put it on top of the jam can."} The original instruction was {\bf "Pick up the shiba dog and put it on top of the jam can."} and there is no shiba dog. This time, instruction is correctly understood by the robot. {\bf ambiguous, success}.
                \end{nolabel_caption}
            \end{center}
        \end{minipage}
        \begin{minipage}[t]{0.23\hsize}
            \begin{center}
                \includegraphics[height=3cm]{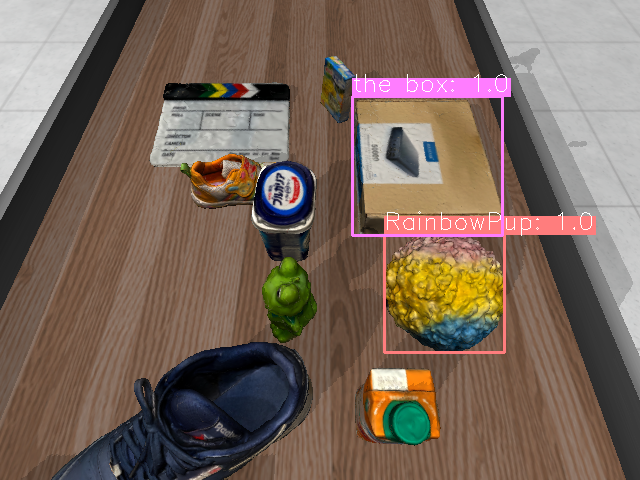}\\
                \vspace{-2mm}
                \begin{nolabel_caption}
                (c) {\bf "Pick up \textbf{\textit{RaimbowPup}} and place it on the box."} The original instruction was {\bf "Pick up the colorful toy and place it on the box."} and the robot picked the green frog toy wrongly. By using the name in instruction, it succeeded. {\bf unambiguous, success}.
                \end{nolabel_caption}
            \end{center}
        \end{minipage}
        \begin{minipage}[t]{0.23\hsize}
            \begin{center}
               \includegraphics[height=3cm]{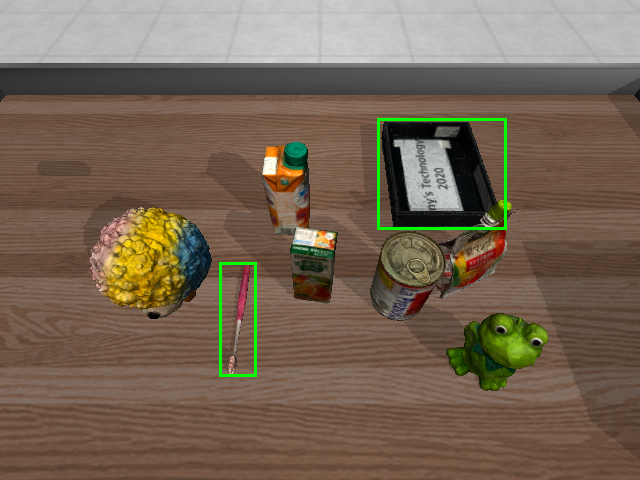}\\
               \vspace{-2mm}
               \begin{nolabel_caption}
               (d) {\bf "Toothbrush needs to be inside \textbf{\textit{Sara}}."} The phrase "A needs to be inside B" was not recognized as pick-and-place instruction. {\bf unambiguous, failure}.
               \end{nolabel_caption}
            \end{center}
        \end{minipage}
    \end{center}
    \vspace{-1mm}
    \caption{{\bf w/ Naming} examples. Green bounding box is ground truth. Pink and orange bounding boxes are predicted {\bf src} and {\bf dst}.}
   \label{fig:result_with_naming}     
\end{figure*}

\cref{fig:result_wthout_naming} shows qualitative results of w/o naming.
It illustrates that how robots fails to disambiguate human instructions.
\cref{fig:result_with_naming} shows qualitative results of w/naming.
The comparisons of \cref{fig:result_wthout_naming} and \cref{fig:result_with_naming} demonstrates that naming objects help robots to solve ambiguity in human instructions.
\vspace{5mm}

\begin{table}[!htb]
    \caption{Real-robot experiment results}
    \begin{center}
        \begin{tabular}{cc}
            \hline
            \multicolumn{2}{c}{\bf{SR} [\%]}  \\
            \hline
            Naming process               & Manipulation process       \\
            95.0  (19/20) & 45.0  (9/20) \\
            \hline
        \end{tabular}
    \end{center}
    \label{table:result_real_robot_experiment}
    \vspace{-3mm}
\end{table}

\begin{figure}[!tb]
    \centering
    \includegraphics[scale=0.24]{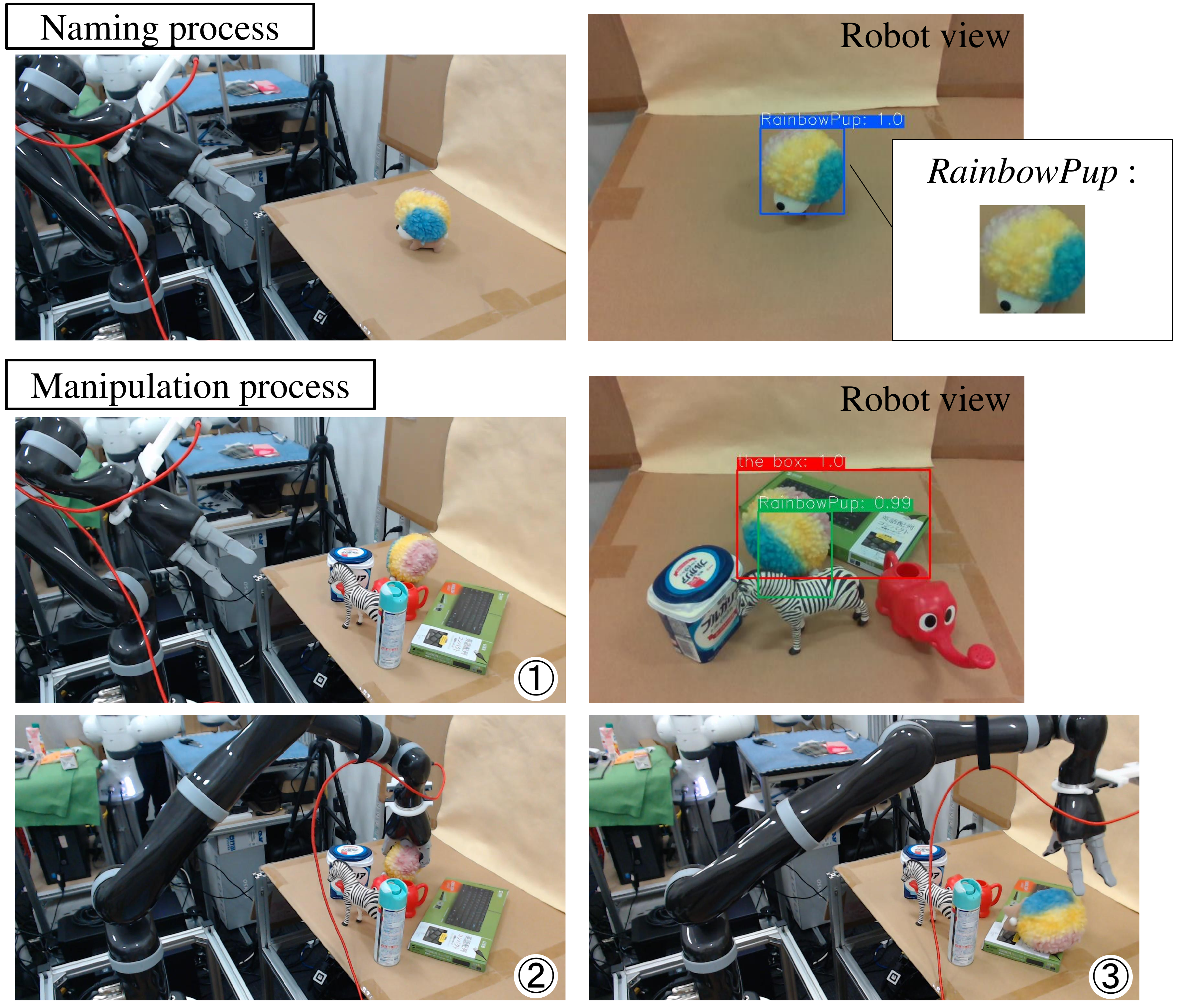}
    \vspace{-5mm}
    \caption{An example of the real-robot experiment results. The naming instruction was \textbf{"The nickname of it is \textbf{\textit{RainbowPup}}."} and the pick-and-place instruction was \textbf{"pick up \textbf{\textit{RainbowPup}} and place it on the box."}.}
    \label{fig:real_robot_env}
\end{figure}

\subsection{Real-Robot Experiment}

We used a robotic manipulator arm, Kinova Jaco2 \cite{kinovarobotics} and a camera, Realsense D435i \cite{d435i} to evaluate our system in real-robot environment.
We tested our method with the real objects that we used in our simulation experiment.
The evaluation procedure is as follows:
\begin{enumerate}
    \item Sample 1 object from the actual objects and place the sampled object on the table.
    \item Input a naming instruction and RGB-D image of the scene to the system.
    \item Sample from 6 to 8 objects including the object we chose at 1) and place them on the table.
    \item Input a pick-and-place instruction including the object's name given in 2) and a RGB-D image of the scene to the system.
    \item Check if the robot achieves the instruction or not.
\end{enumerate}
We asked 1 volunteer to give instructions for each scene, and we ran this procedure 20 times.

Quantitative result of real-robot experiment is shown in \cref{table:result_real_robot_experiment}. 
\cref{fig:real_robot_env} shows an example movement of the real-robot and the prediction of the system. 
From \cref{table:result_real_robot_experiment} and \cref{fig:real_robot_env}, we can confirm that our system works also in the real-robot environment.
However, the success rate was lower than that of simulation.
One of the reasons is the performance degrade in sim2real transfer. 
I-MDETR are trained with the dataset generated in the simulator.
Improving the performance in the real-robot environment is our future work.
The actual robot movements can be also observed in our supplementary video.

\section{CONCLUSIONS}
\label{sec:CONCLUSIONS}

To test our hypothesis that naming target objects in advance mitigates the ambiguity in natural language instructions,
we have built a robot system that can remember the name of an object together with its appearance, and understand instructions using the name.
Our experiments have shown that naming target objects and manipulating them by using the names significantly improves the success rate of manipulation.
Naming is not only for reducing the ambiguity of instruction, but we believe it is also helpful for giving complicated instructions such as motion trajectory control by natural language.
Investigating other positive effects of naming is our future work.
\appendices
\section{Training Dataset Generation for I-MDETR}
\label{appendix:data_gen_IMDETR}

We use a physics simulator, Mujoco \cite{todorov2012mujoco} to generate a training dataset for I-MDETR.
We scanned 60 common household objects by an iOS app, metascan \cite{metascan}.
After constructing collision volumes with Volumetric Hierarchical Approximate Convex Decomposition \cite{VHACD}, we introduced these objects into Mujoco \cite{todorov2012mujoco}. Except for the scanned objects such as tables and floors are based on metaworld \cite{yu2019metaworld} environment. Metaworld is an open-source simulated benchmark for multi-task learning in robotic research.

Our dataset generation is composed of two steps, scene generation and instruction labelling for each scene.
In this research, we assumed that the objects are placed on a table such as \cite{shridhar2022cliport}.
For generating scenes, we first sampled 1 to 10 objects randomly and placed them on a table.
We also sampled camera pose from pre-defined 4 poses where the camera look at the table from the right side, the left side, the front side and the back side, and then rendered the scene.
We generated 10,000 scenes for I-MDETR training dataset.

Referring to the object information of each scene, the instruction generator automatically generated instructions.
The instruction generator randomly selects objects present in the scene and generates instruction text.
The instruction text is generated from the instruction expression templates and the referring expression templates created by a human annotator.
The instruction expression templates are associated with each instruction class, with a total of about 200, and the referring expression templates are associated with each scanned object, with a total number of about 15,000.

The referring expressions of the objects are generated at various levels, from the most detailed (e.g., "hand soap in the white and light purple dispenser") to the simplest (e.g., "it"), and are selected on the conditions that the object is uniquely determined by the expression in relation to the other objects simultaneously present on the table.
For example, if there is only one object on the table, "pick \textbf{it} up" is sufficient to identify the target object, but if there are many objects on the table, then a more detailed expression is necessary.

To generate instructions that refer to objects by their names, we collected variety of names by crawling Web.
We collected person names in several languages, brand names for various products, and nicknames, such as those used for toys.
We generated instructions using the names by stochastically replacing the referring expressions with the collected names in the instructions.

By combining the options for additional expressions (e.g., "will you" "please"), with or without articles, and with or without modification, it is possible to generate a variety of expressions.
Negative instructions (i.e., instructions for non-existent objects, non-executable instructions, scene descriptions that are not instructions, and random sentences) can also be generated by the instruction generator.
Training with these negative instructions, we expect our robot to learn to understand the sentences that do not require actions.
With this procedure, we generated 25 instructions for each scene (total 250,000).
Based on the MDETR weights pre-trained with RefCOCOg \cite{mao2016generation}, we finetuned I-MDETR model with our generated dataset.

\section{Few-shot Object Detection}
\label{appendix:few-shot}
For simulation experiments, we use Unseen Object Instance Segmentation (UOIS) \cite{xie2021uois} and ArcFace \cite{deng2019arcface} combination to realize few-shot object detection.
UOIS is trained on the same generated dataset as I-MDETR.
ArcFace is trained on RGB-D Object Dataset \cite{lai2011large}.
In naming process, the image of the named object is stored in robot's memory, and in pick-and-place instruction, UOIS extracts the regions of all objects on the table and generates an image template for each object.
Each image template is fed to ArcFace along with the image of the named object. These features are put into Support Vector Machine (SVM).
SVM predicts whether the features of each template are classified into the class of the named object or not.
For real-robot experiments, we used MixFormer \cite{cui2022mixformer} trained on GOT10K \cite{huang2019got} for few-shot object detection.  

\section{Target Object Pick-and-Place Method}
\label{appendix:VGN}
Our object picking method is based on Volumetric Grasp Network (VGN) \cite{breyer2021vgn}.
We implemented kinova jaco2 gripper model in Mujoco referring to the original URDF \cite{kinovarobotics}.
After generating a dataset in simulation, we trained the model via supervised learning following the paper \cite{breyer2021vgn}.
VGN is a method for predicting grasp poses for the whole scene so it is required to filter irrelevant grasp poses to grasp the target object.
To realize this, we use UOIS \cite{xie2021uois} as well as our few-shot object detection method for simulation.
The target object segmentation map is obtained from the target object's bounding box and the segmentation map, and then we can filter the grasp candidates.
For placing the object to the target bounding box, we estimate the center position by taking the average of four points positions of the bounding box.

\section*{ACKNOWLEDGMENT}
We would like to thank Wei Jiang and Noel Chen for implementing and evaluating the first versions of I-MDETR during their internship in Sony.
We also would like to thank Hirotaka Suzuki for helpful discussions.

\addtolength{\textheight}{-12cm}   

\bibliographystyle{IEEEtran}
\bibliography{ours}

\end{document}